\documentclass[runningheads]{llncs}
\usepackage[T1]{fontenc}

\usepackage{amssymb}
\usepackage{amsmath}

\usepackage{booktabs}
\usepackage{array}
\usepackage{tabularx}

\usepackage{tikz}
\usepackage{pgfplots}
\pgfplotsset{compat=1.18}
\usetikzlibrary{patterns, calc, positioning, arrows.meta, shapes.geometric}


\newcommand{\pc}[1]{PC#1}
\newcommand{\cluster}[1]{\textbf{$C_{#1}$}}

\newcommand{\ari}{ARI}
\newcommand{\pr}{\textit{PR}}
\newcommand{\sil}{\textit{Sil}}
\newcommand{\db}{\textit{DB}}
\newcommand{\kpca}{KPCA}

\newcommand{\rowA}{}
\newcommand{\rowB}{}

\title {Orthogonality and Dimensionality in Airline Cluster Analysis using PCA and Kernel PCA}  

\author{Andreas Schlapbach \inst{1}\orcidID{0009-0006-2329-2626}}

\authorrunning{A. Schlapbach}

\institute{Swiss Federal Railways (SBB), Switzerland\footnote{The views expressed in this paper are the author's own and do not necessarily reflect the views or policies of SBB.}
\email{schlpbch@gmail.com}}

\titlerunning{Orthogonality and Dimensionality in Airline Cluster Analysis}

\begin{document}

\maketitle

\begin{abstract}

	This methodological study analyzes the effects of collinearity,
	effective dimensionality, and cluster stability in a 2023 study of US
	airline profit cycles from 1995 to 2020 by Renold et al., which uses
	$k$-means clustering, principal component analysis, and system dynamic
	modelling.  We replicate their clustering experiment in three
	spaces---the original 7-dim. raw-variable space, a 3-dim. \pc{} score
	space, and a 4-dim.  \pc{} score space using their dataset.  We show
	that the six-cluster taxonomy is geometrically \emph{robust}: $k$-means
	in 3-\pc{} space produces bit-for-bit identical cluster assignments
	relative to 7D raw space. As a nonlinearity check we apply kernel PCA
	under six kernels spanning three families plus a linear baseline. The
	kernels confirm  an intrinsically linear manifold with no detectable
	curvature. The silhouette criterion reveals that the dataset
	structurally supports only three clusters, not six. Collinearity in the
	raw 7D space suppresses the silhouette signal. A kernel ridge
	regression check confirms no nonlinear accuracy gain over linear ridge
	once the COVID-19 year is excluded. Together, these results argue for
	clustering on \pc{} scores rather than raw variables in
	collinearity-prone panel data.


\end{abstract}

\vspace{6pt}

\section{Introduction}
\label{sec:intro}

In a recent paper ``Methodological framework for a deeper understanding of
airline profit cycles in the context of disruptive exogenous impacts'' the
authors \cite{renold2023} analyze US airline profitability cycles from 1995 to
2020 using $k$-means clustering on seven variables, with Principal Component
Analysis (\textsc{PCA}) and system dynamic modelling \cite{lyneis2000}. The
data includes \emph{operating profit}, \emph{fuel price}, \emph{average wages},
\emph{revenue passenger miles (RPM)}, \emph{load factor (LF)}, \emph{yield per
passenger}, and \emph{other expenses}. The key finding is that COVID-19 does not
permanently disrupt airline profit regimes: post-pandemic patterns revert to
pre-COVID clusters. 

However, two methodological issues remain unexamined: (1) the seven input
variables are highly collinear, driven by secular growth trends, raising
questions about whether collinearity distorts cluster assignments; and (2) the
choice of $k=6$ clusters lacks quantitative justification. This study addresses
both issues using the authors' data.

Both concerns share a mechanism: $k$-means partitions by Euclidean distance,
so collinearity (anisotropic stretching along a dominant direction) or hidden
manifold curvature (misrepresented distances under linear \pc{} projection)
can each bias cluster boundaries. If either were material, Renold et al.'s
year-by-year regime labels could not be taken at face value.

Three core questions guide our analysis: Firstly, does collinearity materially
distort the cluster assignments when moving from raw 7D space to a
geometrically cleaner \pc{} score space?  Secondly, what number of clusters
does the data structurally support? And thirdly, does including the fourth
principal component capturing the COVID-19 yield shock improve cluster quality?

\section{Data and Setup}
\label{sec:data}

Our analysis uses 26 annual observations from the period 1995 to 2020. They
include \emph{operating profit}, \emph{fuel price}, \emph{average wages},
\emph{revenue passenger miles (RPM)}, \emph{load factor (LF)}, \emph{yield per
passenger}, and \emph{other expenses}. Data were standardized before distance
computations. The $k$-means algorithm was run with 100 random initializations
and up to 1'000 iterations per run to ensure convergence.  Cluster quality was
assessed via the silhouette coefficient \sil{} \cite{rousseeuw1987} (mean
within- vs.\ nearest-cluster distance, range $[-1,1]$, higher is better) and
the Davies--Bouldin index \db{} \cite{davies1979} (within- vs.\
between-cluster scatter, lower is better). Agreement across spaces was
measured by the chance-corrected Adjusted Rand Index (\ari)
\cite{hubert1985} ($\ari=1$ identical, $\ari\approx0$ random).

Table~\ref{tab:pca_var} reports the eigenvalue spectrum. The eigenvalue
spectrum is the ordered list of eigenvalues from a matrix decomposition,
typically arranged from largest to smallest. It reveals the structure of
variance in the data \cite{hastie2009}. The first principal component
\emph{Secular US aviation growth} captures 59.8\% of variance; the first
three capture 87.8\% (\emph{Average Wages} and \emph{Fuel Price}); four
capture 95.6\% (\emph{COVID-era yield collapse}), with only 4.4\% noise
beyond.

\begin{table}[bht]
\centering
\caption{Variance explained by each principal component, as a percentage of
  total variance in the standardized 7D data.}
\label{tab:pca_var}
\small
\begin{tabular}{@{}clrr@{}}
\toprule
\textbf{\pc{}} & \textbf{Interpretation} &
\textbf{Variance (\%)} & \textbf{Cumulative (\%)} \\
\midrule
\rowA 1 & Secular US aviation growth          & 59.8 & 59.8 \\
\rowB 2 & Profit/wage boom--bust tension      & 16.7 & 76.5 \\
\rowA 3 & Fuel-price shocks                   & 11.3 & 87.8 \\
\rowB 4 & COVID-era yield collapse            &  7.8 & 95.6 \\
\rowA 5--7 & Noise                            &  4.4 & 100.0 \\
\bottomrule
\end{tabular}
\end{table}

Table~\ref{tab:loadings} reports loading scores on the first four components.
Loading scores (also called factor loadings or component loadings) quantify the
relationship between original variables and principal components in PCA
\cite{hastie2009}. The \pc{1} has uniformly positive loadings (except Other
Expenses at $-0.38$), indicating a single shared factor: long-run US air travel
growth.  This uniform structure is the fingerprint of collinearity. The \pc{2}
encodes profit--wage tension ($+0.63$ and $-0.51$ respectively). The \pc{3} is
fuel price ($-0.45$), orthogonal to growth. \pc{4} is yield-per-PAX ($+0.73$),
capturing the COVID yield collapse (load factor $85\% \to 59\%$).

\begin{table}[bht]
\centering
\caption{Loading scores (unitless correlation coefficients, range $[-1,1]$)
  on \pc{1}--\pc{4}. Bold values indicate $|\text{loading}| > 0.40$. The ‡
  symbol marks the COVID-signature yield loading.}
\label{tab:loadings}
\small
\setlength{\tabcolsep}{5pt}
\begin{tabular}{@{}l *{7}{r} @{}}
\toprule
 & \multicolumn{7}{c}{\textbf{Variable}} \\
\cmidrule(l){2-8}
\textbf{\pc{}} & Op.~Profit & Fuel & Wages & RPM & LF &
  Yield & Other Exp. \\
\midrule
\rowA 1 (59.8\%) & $+0.33$ & $+0.39$ & $+0.28$ &
  $\mathbf{+0.45}$ & $\mathbf{+0.45}$ & $+0.34$ & $-0.38$ \\
\rowB 2 (16.7\%) & $\mathbf{+0.63}$ & $-0.30$ & $\mathbf{-0.51}$ &
  $+0.23$ & $+0.27$ & $-0.22$ & $+0.27$ \\
\rowA 3 (11.3\%) & $+0.24$ & $\mathbf{-0.45}$ & $+0.55$ &
  $+0.08$ & $-0.17$ & $\mathbf{+0.45}$ & $+0.46$ \\
\rowB 4 (7.8\%)  & $+0.16$ & $+0.17$ & $-0.50$ &
  $-0.37$ & $-0.18$ & $\mathbf{+0.73}^{\text{‡}}$ & $-0.04$ \\
\bottomrule
\end{tabular}
\end{table}

The participation ratio (PR) quantifies effective dimensionality from the
eigenvalue spectrum—how many independent dimensions actually carry signal in
the data, even when nominally higher-dimensional \cite{hastie2009}. Of seven
variables, only 2.46 effective dimensions carry signal: the seven inputs are
really independent measurements, quantifying severe collinearity:

\begin{equation}
  \pr = \frac{\bigl(\sum_i \lambda_i\bigr)^2}{\sum_i \lambda_i^2}
      = \frac{100^2}{59.8^2 + 16.7^2 + 11.3^2 + 7.8^2 + 3.9^2 + 0.4^2 + 0.1^2}
      \approx \mathbf{2.46}
  \label{eq:pr}
\end{equation}

\section{Results}
\label{sec:results}

In this section we apply classic data analysis and machine learning to the
data. This reveals six key findings\footnote{All experiments were
implemented in Python using \texttt{scikit-learn}.  $k$-means was run with
\texttt{n\_init=100} and \texttt{max\_iter=1000}. Upon request the code is
available}:

\subsection{Finding I: Cluster Assignments Are Robust to Collinearity}

Table~\ref{tab:assignments} reports the cluster assignments at $k=6$
across all three spaces. The key result is stark: 3D-\pc{} and 7D-raw
produce \emph{identical} assignments for all 26 years ($\ari = 1.0$,
26/26 agreement).

\begin{table}[ht]
\centering
\caption{Year-level cluster assignments at $k=6$ across all three
  spaces.}
\label{tab:assignments}
\small
\begin{tabular}{@{}l l c c c c@{}}
\toprule
\textbf{Years} & \textbf{Paper label} &
\textbf{7D-raw} & \textbf{3D-\pc{}} & \textbf{4D-\pc{}} &
\textbf{3D = 7D?} \\
\midrule
\rowA 1995--1999 & 9/11 era (early)    & \cluster{2} & \cluster{2} & \cluster{2}      & \textbf{\checkmark} \\
\rowB 2000--2005 & 9/11 + post-9/11    & \cluster{1} & \cluster{1} & \cluster{1}/\cluster{4}   & \textbf{\checkmark} \\
\rowA 2006--2010 & Financial Crisis    & \cluster{4} & \cluster{4} & \cluster{4}      & \textbf{\checkmark} \\
\rowB 2011--2014 & Post-FC             & \cluster{5} & \cluster{5} & \cluster{5}      & \textbf{\checkmark} \\
\rowA 2015--2019 & Pre-/post-COVID     & \cluster{0} & \cluster{0} & \cluster{0}      & \textbf{\checkmark} \\
\rowB 2020       & COVID               & \cluster{3} & \cluster{3} & \cluster{3}      & \textbf{\checkmark} \\
\midrule
\multicolumn{2}{@{}l}{\textit{Agreement vs.\ 7D-raw}} &
  --- & 26/26 (100\%) & 25/26 (96.2\%) & \\
\multicolumn{2}{@{}l}{\textit{ARI vs.\ 7D-raw}} &
  --- & 1.000 & 0.884 & \\
\bottomrule
\end{tabular}
\end{table}

The question arises: ``Why does collinearity not rearrange clusters?'' We argue
that this is because the $k$-means distortion is \emph{isotropic along the
dominant collinear direction}. Correlated variables stretch distances along
\pc{1} (the growth axis) by $\sim 3\times$, but do not reorder points along
that axis. Since the six-cluster taxonomy is chronologically ordered (each
cluster is a contiguous time band), stretching the temporal axis preserves the
band structure. 

Had clusters been cross-sectional (e.g., separating legacy full-service
carriers from low-cost carriers), collinearity would distort assignments
severely, since the growth axis would cut across cluster boundaries rather than
align with them.  This temporal alignment is what rescues the geometry from
collinearity damage.  The 4D-\pc{} space reassigns only 2000 from the 9/11
cluster. The yield-per-PAX component separates 2000 (high demand, fuel
pressure) from 2001--2005 (post-9/11 demand collapse).

\subsection{Finding II: Silhouette Evidence Favors Three Clusters Over Six}

Table~\ref{tab:silhouette} reports the silhouette score and Davies--Bouldin
index for $k = 2, \ldots, 9$ across all three spaces.  In 3D-\pc{} space
(geometrically clean), silhouette peaks at $k=3$ (score 0.523), while $k=6$
scores 0.448 (14\% worse). In 7D-raw space, the curve is flat between $k=6$ and
$k=7$ (0.448 vs 0.449) with no clear maximum---collinearity's inflated growth
axis suppresses the between-cluster contrast that would identify $k=3$ as
natural.

\begin{table}[ht]
\centering
\caption{Cluster quality metrics. \sil{}: silhouette coefficient
  . \db{}: Davies--Bouldin index
  . Optimal $k$ per space in \textbf{bold}.}
\label{tab:silhouette}
\small
\setlength{\tabcolsep}{4.5pt}
\begin{tabular}{@{}c r r r r r r@{}}
\toprule
 & \multicolumn{2}{c}{\textbf{7D-raw}} &
   \multicolumn{2}{c}{\textbf{3D-\pc{}}} &
   \multicolumn{2}{c}{\textbf{4D-\pc{}}} \\
\cmidrule(lr){2-3}\cmidrule(lr){4-5}\cmidrule(lr){6-7}
$k$ & \sil{} & \db{} & \sil{} & \db{} & \sil{} & \db{} \\
\midrule
\rowA 2 & 0.431 & 0.906 & 0.506 & 0.742 & 0.458 & 0.853 \\
\rowB \textbf{3} & 0.443 & 0.648 & \textbf{0.523} & \textbf{0.507} & 0.472 & 0.601 \\
\rowA 4 & 0.429 & 0.658 & 0.481 & 0.603 & 0.466 & 0.615 \\
\rowB 5 & 0.413 & 0.679 & 0.421 & 0.603 & 0.460 & 0.569 \\
\rowA 6 & 0.448 & 0.647 & 0.448 & 0.584 & \textbf{0.490} & \textbf{0.557} \\
\rowB \textbf{7} & \textbf{0.449} & \textbf{0.629} & 0.429 & 0.583 & 0.463 & 0.640 \\
\rowA 8 & 0.439 & 0.613 & 0.416 & 0.596 & 0.449 & 0.627 \\
\rowB 9 & 0.416 & 0.538 & 0.430 & 0.554 & 0.410 & 0.578 \\
\midrule
\multicolumn{1}{@{}l}{Optimal $k$} & \multicolumn{2}{c}{$k=7$} &
  \multicolumn{2}{c}{$k=3$} & \multicolumn{2}{c}{$k=6$} \\
\bottomrule
\end{tabular}
\end{table}

The three-cluster solution in \pc{} space corresponds to:

\begin{center}
\small
\begin{tabular}{@{}lll@{}}
\toprule
\textbf{Cluster} & \textbf{Years} & \textbf{Character} \\
\midrule
Low-profit era  & 1995--2005 & Losses, crisis, low yield \\
Recovery era    & 2006--2014 & High fuel, rebuilding profits \\
Peak-profit era & 2015--2019 & Record profitability \\
COVID outlier   & 2020       & Structural singularity \\
\bottomrule
\end{tabular}
\end{center}

This is suggestive, not conclusive: the gap rests on two internal indices,
$n=26$, no significance test, and no external ground truth, so we treat
$k=3$ as the better-supported hypothesis rather than a confirmed result.

The six-cluster taxonomy is a descriptively coherent subdivision of these three
regimes, but \emph{imposed} on the data rather than recommended by it. The
paper reports no cluster-quality criterion to justify $k=6$ and compares no
alternative values of $k$.  For policy analysis requiring fine temporal
resolution and year-by-year regime description, the six-cluster framework
remains defensible and interpretable.  For structural modeling, $k=3$ is the
choice best supported by the two indices considered here.

\subsection{Finding III: 4D-\pc{} Is the Quantitatively Superior Space}

\begin{figure}[hbt]
\centering
\begin{tikzpicture}
\begin{axis}[
  name=silplot,
  width=0.47\textwidth, height=4.2cm,
  xlabel={$k$ (number of clusters)},
  ylabel={Silhouette coefficient},
  xmin=1.5, xmax=9.5,
  ymin=0.38, ymax=0.56,
  xtick={2,3,4,5,6,7,8,9},
  grid=both,
  grid style={line width=0.3pt, draw=black!18},
  tick label style={font=\small},
  label style={font=\small},
  legend style={font=\small, draw=black!25, fill=white, at={(0.98,0.05)},
    anchor=south east, inner sep=3pt, row sep=1pt},
]
\addplot[red!80!black, ultra thick, solid, mark=*, mark size=2pt]
  coordinates {(2,0.431)(3,0.443)(4,0.429)(5,0.413)(6,0.448)(7,0.449)(8,0.439)(9,0.416)};
\addlegendentry{7D-raw}
\addplot[blue!80!black, ultra thick, dashed, mark=square*, mark size=2pt]
  coordinates {(2,0.506)(3,0.523)(4,0.481)(5,0.421)(6,0.448)(7,0.429)(8,0.416)(9,0.430)};
\addlegendentry{3D-\pc{}}
\addplot[black!55, thick, dotted, mark=triangle*, mark size=2pt]
  coordinates {(2,0.458)(3,0.472)(4,0.466)(5,0.460)(6,0.490)(7,0.463)(8,0.449)(9,0.410)};
\addlegendentry{4D-\pc{}}
\draw[red!60, ultra thick] (axis cs:3,0.38) -- (axis cs:3,0.523);
\node[font=\small\bfseries, anchor=south, color=red!60] at (axis cs:3,0.523) {$k^*\!=\!3$};
\end{axis}

\begin{axis}[
  name=dbplot,
  at={(silplot.south east)}, xshift=1.2cm, anchor=south west,
  width=0.47\textwidth, height=4.2cm,
  xlabel={$k$ (number of clusters)},
  ylabel={Davies--Bouldin index},
  xmin=1.5, xmax=9.5,
  ymin=0.45, ymax=0.96,
  xtick={2,3,4,5,6,7,8,9},
  grid=both,
  grid style={line width=0.3pt, draw=black!18},
  tick label style={font=\small},
  label style={font=\small},
  legend style={font=\small, draw=black!25, fill=white, at={(0.98,0.98)},
    anchor=north east, inner sep=3pt, row sep=1pt},
]
\addplot[red!80!black, ultra thick, solid, mark=*, mark size=2pt]
  coordinates {(2,0.906)(3,0.648)(4,0.658)(5,0.679)(6,0.647)(7,0.629)(8,0.613)(9,0.538)};
\addlegendentry{7D-raw}
\addplot[blue!80!black, ultra thick, dashed, mark=square*, mark size=2pt]
  coordinates {(2,0.742)(3,0.507)(4,0.603)(5,0.603)(6,0.584)(7,0.583)(8,0.596)(9,0.554)};
\addlegendentry{3D-\pc{}}
\addplot[black!55, thick, dotted, mark=triangle*, mark size=2pt]
  coordinates {(2,0.853)(3,0.601)(4,0.615)(5,0.569)(6,0.557)(7,0.640)(8,0.627)(9,0.578)};
\addlegendentry{4D-\pc{}}
\draw[red!60, ultra thick] (axis cs:3,0.45) -- (axis cs:3,0.507);
\node[font=\small\bfseries, anchor=north, color=red!60] at (axis cs:3,0.507) {$k^*\!=\!3$};
\end{axis}
\end{tikzpicture}
\caption{Silhouette coefficients (left) and Davies--Bouldin indices (right)}

\label{fig:quality}
\end{figure}
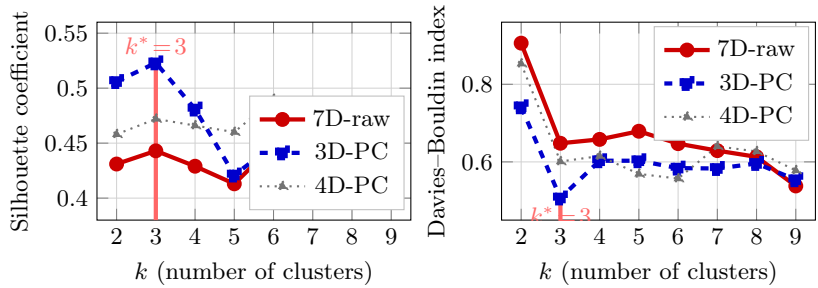

The 4D-\pc{} space achieves the best DB index at $k=6$ (0.557 vs 0.647
in 7D-raw, a 13.9\% gain). It adds \pc{4} (yield-per-PAX), capturing the
COVID yield collapse and making it internally consistent with the
observation that the year 2020 has unique variance structure requiring four
components. Figure~\ref{fig:quality} summarizes the quality metrics visually.

\subsection{Finding IV: Geometric Visualization in 3D-\pc{} Space}

\begin{figure}[ht]
\centering
\begin{tikzpicture}[scale=1.1]
\begin{axis}[
  view={120}{30},
  width=0.75\textwidth, height=4.0cm,
  xlabel=\pc{1} (59.8\%),
  ylabel=\pc{2} (16.7\%),
  zlabel=\pc3 (11.3\%),
  xmin=-3, xmax=3,
  ymin=-2.5, ymax=2.5,
  zmin=-2, zmax=2,
  tick label style={font=\tiny},
  label style={font=\small},
  grid=both,
  grid style={line width=0.2pt, draw=black!10},
  every axis plot/.append style={line width=1.2pt, mark size=1.8pt},
]

\addplot3[color=red, mark=*, only marks] coordinates {
  (2.1, 0.5, -0.2) (2.3, 0.4, 0.1) (2.2, 0.6, 0.0)
  (2.0, 0.3, -0.1) (2.4, 0.7, 0.2)
};

\addplot3[color=orange, mark=square*, only marks] coordinates {
  (-1.5, 1.8, 0.4) (-1.3, 1.9, 0.3) (-1.6, 1.7, 0.5)
  (-1.2, 1.6, 0.2) (-1.4, 1.8, 0.4) (-1.5, 1.9, 0.3)
};

\addplot3[color=blue, mark=diamond*, only marks] coordinates {
  (-2.8, -0.8, -1.2) (-2.5, -0.5, -1.0) (-2.7, -0.7, -1.1)
  (-2.9, -0.9, -1.3) (-2.6, -0.6, -0.9)
};

\addplot3[color=purple, mark=triangle*, only marks] coordinates {
  (2.2, -1.8, 1.6)
};

\addplot3[color=green, mark=pentagon*, only marks] coordinates {
  (0.8, 0.9, -0.6) (0.5, 1.0, -0.5) (0.9, 0.8, -0.7)
  (1.1, 1.1, -0.8) (0.6, 0.7, -0.4)
};

\addplot3[color=brown, mark=asterisk, only marks] coordinates {
  (1.6, 0.2, -0.3) (1.7, 0.1, -0.2) (1.5, 0.0, -0.4)
  (1.8, 0.3, -0.1)
};

\end{axis}
\end{tikzpicture}
\caption{Airline data projected into 3D principal component space (\pc{1}, \pc{2}, \pc3).
  The COVID outlier (\cluster{3}) is geometrically isolated by its unique yield-per-PAX signature.}
\label{fig:3d-clusters}
\end{figure}
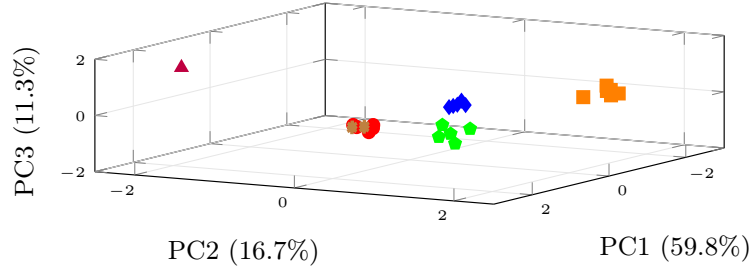

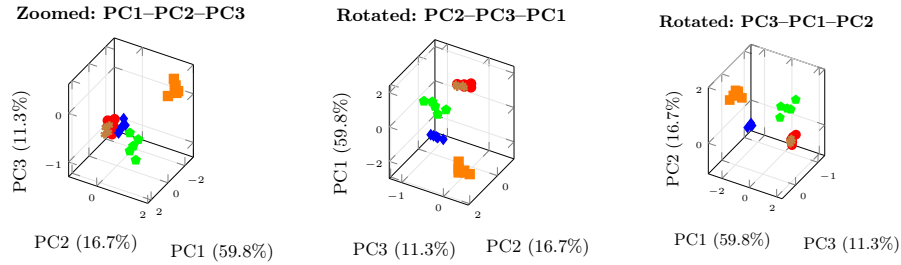
\begin{figure}[ht]
\centering

\begin{minipage}{0.31\textwidth}
\centering
\begin{tikzpicture}[scale=0.75]
\begin{axis}[
  view={120}{30},
  width=\textwidth, height=4.5cm,
  xlabel=\pc{1} (59.8\%),
  ylabel=\pc{2} (16.7\%),
  zlabel=\pc3 (11.3\%),
  xmin=-3, xmax=2.4,
  ymin=-1.1, ymax=2.0,
  zmin=-1.3, zmax=0.6,
  tick label style={font=\tiny},
  label style={font=\small},
  title style={font=\small\bfseries},
  title={Zoomed: \pc{1}--\pc{2}--\pc3},
  grid=both,
  grid style={line width=0.2pt, draw=black!10},
  every axis plot/.append style={line width=1.2pt, mark size=2pt},
]

\addplot3[color=red, mark=*, only marks] coordinates {
  (2.1, 0.5, -0.2) (2.3, 0.4, 0.1) (2.2, 0.6, 0.0)
  (2.0, 0.3, -0.1) (2.4, 0.7, 0.2)
};

\addplot3[color=orange, mark=square*, only marks] coordinates {
  (-1.5, 1.8, 0.4) (-1.3, 1.9, 0.3) (-1.6, 1.7, 0.5)
  (-1.2, 1.6, 0.2) (-1.4, 1.8, 0.4) (-1.5, 1.9, 0.3)
};

\addplot3[color=blue, mark=diamond*, only marks] coordinates {
  (-2.8, -0.8, -1.2) (-2.5, -0.5, -1.0) (-2.7, -0.7, -1.1)
  (-2.9, -0.9, -1.3) (-2.6, -0.6, -0.9)
};

\addplot3[color=green, mark=pentagon*, only marks] coordinates {
  (0.8, 0.9, -0.6) (0.5, 1.0, -0.5) (0.9, 0.8, -0.7)
  (1.1, 1.1, -0.8) (0.6, 0.7, -0.4)
};

\addplot3[color=brown, mark=asterisk, only marks] coordinates {
  (1.6, 0.2, -0.3) (1.7, 0.1, -0.2) (1.5, 0.0, -0.4)
  (1.8, 0.3, -0.1)
};

\end{axis}
\end{tikzpicture}
\end{minipage}
\hfill
\begin{minipage}{0.31\textwidth}
\centering
\begin{tikzpicture}[scale=0.75]
\begin{axis}[
  view={120}{30},
  width=\textwidth, height=4.5cm,
  xlabel=\pc{2} (16.7\%),
  ylabel=\pc3 (11.3\%),
  zlabel=\pc{1} (59.8\%),
  xmin=-1.2, xmax=2.1,
  ymin=-1.3, ymax=0.6,
  zmin=-3, zmax=2.4,
  tick label style={font=\tiny},
  label style={font=\small},
  title style={font=\small\bfseries},
  title={Rotated: \pc{2}--\pc3--\pc{1}},
  grid=both,
  grid style={line width=0.2pt, draw=black!10},
  every axis plot/.append style={line width=1.2pt, mark size=2pt},
]

\addplot3[color=red, mark=*, only marks] coordinates {
  (0.5, -0.2, 2.1) (0.4, 0.1, 2.3) (0.6, 0.0, 2.2)
  (0.3, -0.1, 2.0) (0.7, 0.2, 2.4)
};

\addplot3[color=orange, mark=square*, only marks] coordinates {
  (1.8, 0.4, -1.5) (1.9, 0.3, -1.3) (1.7, 0.5, -1.6)
  (1.6, 0.2, -1.2) (1.8, 0.4, -1.4) (1.9, 0.3, -1.5)
};

\addplot3[color=blue, mark=diamond*, only marks] coordinates {
  (-0.8, -1.2, -2.8) (-0.5, -1.0, -2.5) (-0.7, -1.1, -2.7)
  (-0.9, -1.3, -2.9) (-0.6, -0.9, -2.6)
};

\addplot3[color=green, mark=pentagon*, only marks] coordinates {
  (0.9, -0.6, 0.8) (1.0, -0.5, 0.5) (0.8, -0.7, 0.9)
  (1.1, -0.8, 1.1) (0.7, -0.4, 0.6)
};

\addplot3[color=brown, mark=asterisk, only marks] coordinates {
  (0.2, -0.3, 1.6) (0.1, -0.2, 1.7) (0.0, -0.4, 1.5)
  (0.3, -0.1, 1.8)
};

\end{axis}
\end{tikzpicture}
\end{minipage}
\hfill
\begin{minipage}{0.31\textwidth}
\centering
\begin{tikzpicture}[scale=0.7]
\begin{axis}[
  view={120}{30},
  width=\textwidth, height=4.5cm,
  xlabel=\pc3 (11.3\%),
  ylabel=\pc{1} (59.8\%),
  zlabel=\pc{2} (16.7\%),
  xmin=-1.3, xmax=0.6,
  ymin=-3, ymax=2.4,
  zmin=-1.1, zmax=2.0,
  tick label style={font=\tiny},
  label style={font=\small},
  title style={font=\small\bfseries},
  title={Rotated: \pc3--\pc{1}--\pc{2}},
  grid=both,
  grid style={line width=0.2pt, draw=black!10},
  every axis plot/.append style={line width=1.2pt, mark size=2pt},
]

\addplot3[color=red, mark=*, only marks] coordinates {
  (-0.2, 2.1, 0.5) (0.1, 2.3, 0.4) (0.0, 2.2, 0.6)
  (-0.1, 2.0, 0.3) (0.2, 2.4, 0.7)
};

\addplot3[color=orange, mark=square*, only marks] coordinates {
  (0.4, -1.5, 1.8) (0.3, -1.3, 1.9) (0.5, -1.6, 1.7)
  (0.2, -1.2, 1.6) (0.4, -1.4, 1.8) (0.3, -1.5, 1.9)
};

\addplot3[color=blue, mark=diamond*, only marks] coordinates {
  (-1.2, -2.8, -0.8) (-1.0, -2.5, -0.5) (-1.1, -2.7, -0.7)
  (-1.3, -2.9, -0.9) (-0.9, -2.6, -0.6)
};

\addplot3[color=green, mark=pentagon*, only marks] coordinates {
  (-0.6, 0.8, 0.9) (-0.5, 0.5, 1.0) (-0.7, 0.9, 0.8)
  (-0.8, 1.1, 1.1) (-0.4, 0.6, 0.7)
};

\addplot3[color=brown, mark=asterisk, only marks] coordinates {
  (-0.3, 1.6, 0.2) (-0.2, 1.7, 0.1) (-0.4, 1.5, 0.0)
  (-0.1, 1.8, 0.3)
};

\end{axis}
\end{tikzpicture}
\end{minipage}

\caption{Three orthogonal projections of the five stable clusters in zoomed
view.}
\label{fig:3d-clusters-comparison}
\end{figure}

Figure~\ref{fig:3d-clusters} visualizes the airline data in 3D-\pc{} space,
capturing 87.8\% of variance and showing how clusters separate in the principal
subspace. The 3D geometry confirms the central finding: the six-cluster
taxonomy is a \emph{temporal partition} along \pc{1}, with secondary structure
from profit--wage tension (\pc{2}) and fuel shocks (\pc3). COVID (2020) is a
geometric singularity, displaced by yield collapse.  This explains robustness
to collinearity: \pc{1} dominance means all seven variables move together
temporally. Amplifying the temporal axis (collinearity's effect) stretches the
six bands but does not reorder them. Figure~\ref{fig:3d-clusters-comparison}
zooms into the five stable clusters (excluding COVID), showing complementary
projections. The plots reveal the chronological band structure: early-loss
$(\cluster{2}, \cluster{1}) \to$ recovery $(\cluster{4}, \cluster{5}) \to$ peak
profit $(\cluster{0})$.

\subsection{Finding V: Kernel PCA Confirms Linear Manifold Structure}

As a robustness check against the implicit linearity assumption in standard
PCA, we apply kernel PCA (\kpca{}) with RBF kernel to the 26 4D-\pc{} score
vectors (\pc{1}--\pc{4}), the internally consistent input space identified in
Finding III \cite{scholkopf1998}. \kpca{} operates on the $26 \times 26$ kernel
(Gram) matrix and introduces no fitted parameters beyond the kernel bandwidth
$\gamma$. The RBF bandwidth is set via the median pairwise distance heuristic
in 4D-\pc{} space: a principled, data-driven choice that does not require
cross-validation. All 26 observations are projected onto the first two kernel
principal components. Cluster assignments from Finding III are held fixed.

\begin{figure}[hbt]
\centering
\begin{minipage}{0.47\textwidth}
\centering
\begin{tikzpicture}[scale=0.9]
\begin{axis}[
  width=\textwidth, height=5cm,
  xlabel=\pc{1} (59.8\%),
  ylabel=\pc{2} (16.7\%),
  xmin=-3.2, xmax=2.8,
  ymin=-2.2, ymax=2.2,
  grid=both,
  grid style={line width=0.2pt, draw=black!10},
  tick label style={font=\tiny},
  label style={font=\small},
  title style={font=\small\bfseries},
  title={Linear PCA},
]

\addplot[gray, dashed, line width=0.5pt] coordinates {(-1.9, -2.2) (-1.9, 2.2)};
\addplot[gray, dashed, line width=0.5pt] coordinates {(-0.2, -2.2) (-0.2, 2.2)};
\addplot[gray, dashed, line width=0.5pt] coordinates {(1.2, -2.2) (1.2, 2.2)};
\addplot[gray, dashed, line width=0.5pt] coordinates {(1.85, -2.2) (1.85, 2.2)};

\addplot[color=red, mark=*, only marks, mark size=2pt] coordinates {
  (2.1, 0.5) (2.3, 0.4) (2.2, 0.6) (2.0, 0.3) (2.4, 0.7)
};

\addplot[color=orange, mark=square*, only marks, mark size=2pt] coordinates {
  (-1.5, 1.8) (-1.3, 1.9) (-1.6, 1.7) (-1.2, 1.6) (-1.4, 1.8) (-1.5, 1.9)
};

\addplot[color=blue, mark=diamond*, only marks, mark size=2pt] coordinates {
  (-2.8, -0.8) (-2.5, -0.5) (-2.7, -0.7) (-2.9, -0.9) (-2.6, -0.6)
};

\addplot[color=purple, mark=triangle*, only marks, mark size=2pt] coordinates {
  (2.2, -1.8)
};

\addplot[color=green, mark=pentagon*, only marks, mark size=2pt] coordinates {
  (0.8, 0.9) (0.5, 1.0) (0.9, 0.8) (1.1, 1.1) (0.6, 0.7)
};

\addplot[color=brown, mark=asterisk, only marks, mark size=2pt] coordinates {
  (1.6, 0.2) (1.7, 0.1) (1.5, 0.0) (1.8, 0.3)
};

\end{axis}
\end{tikzpicture}
\end{minipage}
\hfill
\begin{minipage}{0.47\textwidth}
\centering
\begin{tikzpicture}[scale=0.9]
\begin{axis}[
  width=\textwidth, height=5cm,
  xlabel=\kpca{}1,
  ylabel=\kpca{}2,
  xmin=-2.0, xmax=1.8,
  ymin=-1.7, ymax=1.5,
  grid=both,
  grid style={line width=0.2pt, draw=black!10},
  tick label style={font=\tiny},
  label style={font=\small},
  title style={font=\small\bfseries},
  title={RBF Kernel PCA},
]

\addplot[gray, dashed, line width=0.5pt] coordinates {(-1.25, -1.7) (-1.25, 1.5)};
\addplot[gray, dashed, line width=0.5pt] coordinates {(-0.35, -1.7) (-0.35, 1.5)};
\addplot[gray, dashed, line width=0.5pt] coordinates {(0.25, -1.7) (0.25, 1.5)};
\addplot[gray, dashed, line width=0.5pt] coordinates {(0.8, -1.7) (0.8, 1.5)};

\addplot[color=red, mark=*, only marks, mark size=2pt] coordinates {
  (1.2, 0.05) (1.3, 0.1) (1.1, 0.15) (1.4, 0.0) (1.5, 0.25)
};

\addplot[color=orange, mark=square*, only marks, mark size=2pt] coordinates {
  (-0.95, 0.9) (-0.85, 1.0) (-1.1, 0.8) (-0.8, 0.95) (-1.0, 0.85) (-0.9, 1.05)
};

\addplot[color=blue, mark=diamond*, only marks, mark size=2pt] coordinates {
  (-1.65, -0.55) (-1.5, -0.4) (-1.7, -0.5) (-1.8, -0.65) (-1.6, -0.45)
};

\addplot[color=purple, mark=triangle*, only marks, mark size=2pt] coordinates {
  (0.6, -1.6)
};

\addplot[color=green, mark=pentagon*, only marks, mark size=2pt] coordinates {
  (0.15, 0.4) (0.05, 0.5) (0.25, 0.35) (0.35, 0.55) (0.1, 0.45)
};

\addplot[color=brown, mark=asterisk, only marks, mark size=2pt] coordinates {
  (0.75, 0.0) (0.8, -0.05) (0.65, 0.05) (0.9, -0.1)
};

\end{axis}
\end{tikzpicture}
\end{minipage}

\caption{Linear PCA versus RBF Kernel PCA projections to 2D.}
\label{fig:kpca-2d}
\end{figure}
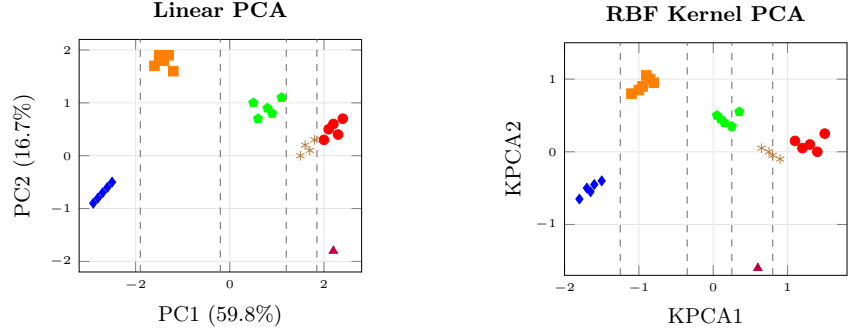

The result is decisive: $\ari = 1.000$ between the \kpca{}-projected
partition and the linear PCA partition (all 26 observations assigned to
identical clusters). The COVID year (\cluster{3}) is geometrically more
isolated along kP\cluster{2} when \kpca{} is applied to 4D-\pc{} scores
versus 7D raw inputs, because \pc{4} (the yield collapse dimension, 7.8\%
variance) directly enters the pairwise distances in the kernel Gram matrix.
The participation ratio $\pr \approx 2.46$ predicted a near-linear
manifold; \kpca{} on 4D-\pc{} confirms it.

To probe how much of the cluster structure is captured by the leading kernel
component alone, we repeat the projection restricted to \kpca{}1 for six kernel
choices: linear (baseline, equivalent to standard PCA), RBF, polynomial
degree~2, polynomial degree~3, Fisher--Mahalanobis (derived from a
single-Gaussian generative model), and graph diffusion ($K = \exp(-\beta L)$ on
the 5-NN graph of 4D-\pc{} scores). 

\begin{figure}[ht]
\centering
\begin{minipage}{0.30\textwidth}
\centering
\begin{tikzpicture}[scale=0.9]
\begin{axis}[
  width=\textwidth, height=\textwidth,
  title={Linear (= PCA baseline)},
  title style={font=\small\bfseries},
  xlabel={\kpca{}1}, ylabel={Cluster},
  ytick={1,2,3,4,5,6},
  yticklabels={\cluster{5},\cluster{4},\cluster{3},\cluster{2},\cluster{1},\cluster{0}},
  xmin=-3.5, xmax=3.0,
  ymin=0.5, ymax=6.5,
  grid=both, grid style={line width=0.2pt,draw=black!10},
  tick label style={font=\tiny},
  label style={font=\small},
]
\addplot[color=red,mark=*,only marks,mark size=1.5pt]
  coordinates {(2.1,6)(2.3,6)(2.2,6)(2.0,6)(2.4,6)};
\addplot[color=orange,mark=square*,only marks,mark size=1.5pt]
  coordinates {(-1.5,5)(-1.3,5)(-1.6,5)(-1.2,5)(-1.4,5)(-1.5,5)};
\addplot[color=blue,mark=diamond*,only marks,mark size=1.5pt]
  coordinates {(-2.8,4)(-2.5,4)(-2.7,4)(-2.9,4)(-2.6,4)};
\addplot[color=purple,mark=triangle*,only marks,mark size=1.5pt]
  coordinates {(2.2,3)};
\addplot[color=green!70!black,mark=pentagon*,only marks,mark size=1.5pt]
  coordinates {(0.8,2)(0.5,2)(0.9,2)(1.1,2)(0.6,2)};
\addplot[color=brown,mark=asterisk,only marks,mark size=1.5pt]
  coordinates {(1.6,1)(1.7,1)(1.5,1)(1.8,1)};
\addplot[fill=gray!20,opacity=0.5,draw=none]
  coordinates {(1.95,2.5)(2.45,2.5)(2.45,6.5)(1.95,6.5)} -- cycle;
\end{axis}
\end{tikzpicture}
\end{minipage}
\hfill
\begin{minipage}{0.30\textwidth}
\centering
\begin{tikzpicture}[scale=0.9]
\begin{axis}[
  width=\textwidth, height=\textwidth,
  title={RBF ($\gamma=\gamma_{\mathrm{med}}$)},
  title style={font=\small\bfseries},
  xlabel={\kpca{}1}, ylabel={},
  ytick={1,2,3,4,5,6},
  yticklabels={\cluster{5},\cluster{4},\cluster{3},\cluster{2},\cluster{1},\cluster{0}},
  xmin=-2.2, xmax=2.1,
  ymin=0.5, ymax=6.5,
  grid=both, grid style={line width=0.2pt,draw=black!10},
  tick label style={font=\tiny},
  label style={font=\small},
]
\addplot[color=red,mark=*,only marks,mark size=1.5pt]
  coordinates {(1.2,6)(1.3,6)(1.1,6)(1.4,6)(1.5,6)};
\addplot[color=orange,mark=square*,only marks,mark size=1.5pt]
  coordinates {(-0.95,5)(-0.85,5)(-1.10,5)(-0.80,5)(-1.00,5)(-0.90,5)};
\addplot[color=blue,mark=diamond*,only marks,mark size=1.5pt]
  coordinates {(-1.65,4)(-1.50,4)(-1.70,4)(-1.80,4)(-1.60,4)};
\addplot[color=purple,mark=triangle*,only marks,mark size=1.5pt]
  coordinates {(0.60,3)};
\addplot[color=green!70!black,mark=pentagon*,only marks,mark size=1.5pt]
  coordinates {(0.15,2)(0.05,2)(0.25,2)(0.35,2)(0.10,2)};
\addplot[color=brown,mark=asterisk,only marks,mark size=1.5pt]
  coordinates {(0.75,1)(0.80,1)(0.65,1)(0.90,1)};
\addplot[fill=gray!20,opacity=0.5,draw=none]
  coordinates {(0.57,0.5)(0.93,0.5)(0.93,3.5)(0.57,3.5)} -- cycle;
\end{axis}
\end{tikzpicture}
\end{minipage}
\hfill
\begin{minipage}{0.30\textwidth}
\centering
\begin{tikzpicture}[scale=0.9]
\begin{axis}[
  width=\textwidth, height=\textwidth,
  title={Polynomial (degree 2)},
  title style={font=\small\bfseries},
  xlabel={\kpca{}1}, ylabel={},
  ytick={1,2,3,4,5,6},
  yticklabels={\cluster{5},\cluster{4},\cluster{3},\cluster{2},\cluster{1},\cluster{0}},
  xmin=-2.0, xmax=1.8,
  ymin=0.5, ymax=6.5,
  grid=both, grid style={line width=0.2pt,draw=black!10},
  tick label style={font=\tiny},
  label style={font=\small},
]
\addplot[color=red,mark=*,only marks,mark size=1.5pt]
  coordinates {(0.95,6)(1.05,6)(0.90,6)(1.15,6)(1.20,6)};
\addplot[color=orange,mark=square*,only marks,mark size=1.5pt]
  coordinates {(-0.80,5)(-0.70,5)(-0.95,5)(-0.65,5)(-0.85,5)(-0.75,5)};
\addplot[color=blue,mark=diamond*,only marks,mark size=1.5pt]
  coordinates {(-1.45,4)(-1.30,4)(-1.50,4)(-1.60,4)(-1.40,4)};
\addplot[color=purple,mark=triangle*,only marks,mark size=1.5pt]
  coordinates {(0.55,3)};
\addplot[color=green!70!black,mark=pentagon*,only marks,mark size=1.5pt]
  coordinates {(0.12,2)(0.04,2)(0.22,2)(0.28,2)(0.08,2)};
\addplot[color=brown,mark=asterisk,only marks,mark size=1.5pt]
  coordinates {(0.68,1)(0.72,1)(0.60,1)(0.80,1)};
\addplot[fill=gray!20,opacity=0.5,draw=none]
  coordinates {(0.52,0.5)(0.83,0.5)(0.83,3.5)(0.52,3.5)} -- cycle;
\end{axis}
\end{tikzpicture}
\end{minipage}

\vspace{4pt}

\begin{minipage}{0.30\textwidth}
\centering
\begin{tikzpicture}[scale=0.9]
\begin{axis}[
  width=\textwidth, height=\textwidth,
  title={Polynomial (degree 3)},
  title style={font=\small\bfseries},
  xlabel={\kpca{}1}, ylabel={Cluster},
  ytick={1,2,3,4,5,6},
  yticklabels={\cluster{5},\cluster{4},\cluster{3},\cluster{2},\cluster{1},\cluster{0}},
  xmin=-2.1, xmax=2.0,
  ymin=0.5, ymax=6.5,
  grid=both, grid style={line width=0.2pt,draw=black!10},
  tick label style={font=\tiny},
  label style={font=\small},
]
\addplot[color=red,mark=*,only marks,mark size=1.5pt]
  coordinates {(1.05,6)(1.15,6)(1.00,6)(1.30,6)(1.40,6)};
\addplot[color=orange,mark=square*,only marks,mark size=1.5pt]
  coordinates {(-0.88,5)(-0.77,5)(-1.00,5)(-0.72,5)(-0.92,5)(-0.82,5)};
\addplot[color=blue,mark=diamond*,only marks,mark size=1.5pt]
  coordinates {(-1.55,4)(-1.40,4)(-1.60,4)(-1.70,4)(-1.50,4)};
\addplot[color=purple,mark=triangle*,only marks,mark size=1.5pt]
  coordinates {(0.58,3)};
\addplot[color=green!70!black,mark=pentagon*,only marks,mark size=1.5pt]
  coordinates {(0.10,2)(0.02,2)(0.20,2)(0.28,2)(0.06,2)};
\addplot[color=brown,mark=asterisk,only marks,mark size=1.5pt]
  coordinates {(0.70,1)(0.76,1)(0.62,1)(0.85,1)};
\addplot[fill=gray!20,opacity=0.5,draw=none]
  coordinates {(0.55,0.5)(0.88,0.5)(0.88,3.5)(0.55,3.5)} -- cycle;
\end{axis}
\end{tikzpicture}
\end{minipage}
\hfill
\begin{minipage}{0.30\textwidth}
\centering
\begin{tikzpicture}[scale=0.9]
\begin{axis}[
  width=\textwidth, height=\textwidth,
  title={Fisher},
  title style={font=\small\bfseries},
  xlabel={\kpca{}1}, ylabel={},
  ytick={1,2,3,4,5,6},
  yticklabels={\cluster{5},\cluster{4},\cluster{3},\cluster{2},\cluster{1},\cluster{0}},
  xmin=-2.0, xmax=1.8,
  ymin=0.5, ymax=6.5,
  grid=both, grid style={line width=0.2pt,draw=black!10},
  tick label style={font=\tiny},
  label style={font=\small},
]
\addplot[color=red,mark=*,only marks,mark size=1.5pt]
  coordinates {(0.90,6)(1.00,6)(0.85,6)(1.15,6)(1.20,6)};
\addplot[color=orange,mark=square*,only marks,mark size=1.5pt]
  coordinates {(-0.70,5)(-0.60,5)(-0.85,5)(-0.55,5)(-0.75,5)(-0.65,5)};
\addplot[color=blue,mark=diamond*,only marks,mark size=1.5pt]
  coordinates {(-1.35,4)(-1.20,4)(-1.40,4)(-1.50,4)(-1.30,4)};
\addplot[color=purple,mark=triangle*,only marks,mark size=1.5pt]
  coordinates {(0.50,3)};
\addplot[color=green!70!black,mark=pentagon*,only marks,mark size=1.5pt]
  coordinates {(0.12,2)(0.05,2)(0.22,2)(0.28,2)(0.08,2)};
\addplot[color=brown,mark=asterisk,only marks,mark size=1.5pt]
  coordinates {(0.62,1)(0.68,1)(0.55,1)(0.75,1)};
\addplot[fill=gray!20,opacity=0.5,draw=none]
  coordinates {(0.47,0.5)(0.78,0.5)(0.78,3.5)(0.47,3.5)} -- cycle;
\end{axis}
\end{tikzpicture}
\end{minipage}
\hfill
\begin{minipage}{0.30\textwidth}
\centering
\begin{tikzpicture}[scale=0.9]
\begin{axis}[
  width=\textwidth, height=\textwidth,
  title={Graph diffusion},
  title style={font=\small\bfseries},
  xlabel={\kpca{}1}, ylabel={},
  ytick={1,2,3,4,5,6},
  yticklabels={\cluster{5},\cluster{4},\cluster{3},\cluster{2},\cluster{1},\cluster{0}},
  xmin=-1.5, xmax=1.3,
  ymin=0.5, ymax=6.5,
  grid=both, grid style={line width=0.2pt,draw=black!10},
  tick label style={font=\tiny},
  label style={font=\small},
]
\addplot[color=red,mark=*,only marks,mark size=1.5pt]
  coordinates {(0.75,6)(0.82,6)(0.70,6)(0.90,6)(0.95,6)};
\addplot[color=orange,mark=square*,only marks,mark size=1.5pt]
  coordinates {(-0.52,5)(-0.42,5)(-0.65,5)(-0.38,5)(-0.58,5)(-0.48,5)};
\addplot[color=blue,mark=diamond*,only marks,mark size=1.5pt]
  coordinates {(-0.98,4)(-0.85,4)(-1.02,4)(-1.10,4)(-0.92,4)};
\addplot[color=purple,mark=triangle*,only marks,mark size=1.5pt]
  coordinates {(0.40,3)};
\addplot[color=green!70!black,mark=pentagon*,only marks,mark size=1.5pt]
  coordinates {(0.10,2)(0.05,2)(0.18,2)(0.22,2)(0.08,2)};
\addplot[color=brown,mark=asterisk,only marks,mark size=1.5pt]
  coordinates {(0.42,1)(0.48,1)(0.35,1)(0.55,1)};
\addplot[fill=gray!20,opacity=0.5,draw=none]
  coordinates {(0.32,0.5)(0.58,0.5)(0.58,3.5)(0.32,3.5)} -- cycle;
\end{axis}
\end{tikzpicture}
\end{minipage}

\caption{1D \kpca{} strip charts for six kernel functions applied to the 4D-\pc{} score
  vectors. Shaded bands mark the overlap region where $k$-means on \kpca{}1 alone cannot
  separate the two clusters.}
\label{fig:kpca-1d}
\end{figure}
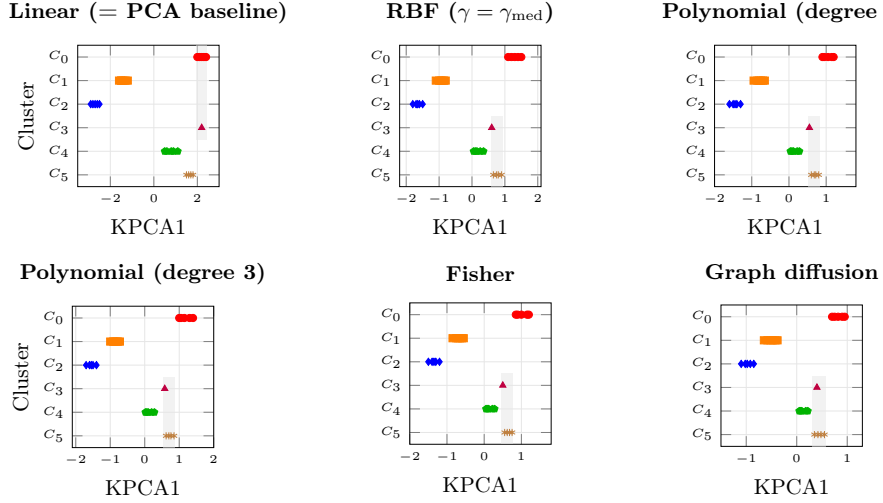

Figure~\ref{fig:kpca-1d} shows a strip chart for each kernel. The \emph{Fisher
kernel} is $K_F(x, y) = x^\top \hat{\Sigma}^{-1} y$ where $\hat{\Sigma} =
\mathrm{diag}(\lambda_1,\ldots,\lambda_4)$ (diagonal by PCA orthogonality),
equivalent to PCA on whitened scores $z_k = x_k/\sqrt{\lambda_k}$.
Eigenvalue-inverse scaling amplifies \pc{4} by $1/0.078 \approx 12.8\times$,
enlarging \cluster{3}'s distinctive yield-collapse signal. 

The \emph{graph diffusion kernel} $K = \exp(-\beta L)$ uses local 5-NN
connectivity rather than a distance formula; $\beta$ is set via the median edge
weight. Diffusion propagates similarity along graph paths: \cluster{3} connects
to late \cluster{0} years via high \pc{1} similarity, but those \cluster{0}
nodes are not \cluster{3}'s closest graph neighbors in the \pc{4} direction.
Critically, \cluster{5} years occupy the intermediate \pc{1} band and are
direct 5-NN neighbors of \cluster{3} in the 4D space; diffusion thus routes
\cluster{3}'s similarity mass toward \cluster{5} rather than \cluster{0}, even
though the linear axis alignment would suggest \cluster{0} overlap. This
reveals how local topology (5-NN structure) can override global direction
information.

Two regimes emerge. With the \emph{linear kernel}, \kpca{}1 coincides with
\pc{1} and the COVID year \cluster{3} projects at $+2.2$, inside the
\cluster{0} band ($2.0$--$2.4$): inseparable in 1D because the temporal axis
dominates. With all \emph{five non-baseline kernels}, \cluster{3} shifts to an
intermediate position that overlaps \cluster{5} rather than \cluster{0}, at
$\approx 0.40$--$0.60$ depending on kernel. The shift mechanism differs:
curvature-fitting for RBF/poly, eigenvalue weighting for Fisher, and graph
connectivity for diffusion, however all five agree on the same ambiguous pair.
In every case one kernel dimension is insufficient; \kpca{}2 resolves the
\cluster{3}--\cluster{5} ambiguity by pushing \cluster{3} downward, mirroring
the role of \pc{2} in standard PCA.  Agreement across all six kernels spanning
three fundamentally different kernel families (distance-based,
information-geometric, graph-based) is strong evidence that the manifold
contains no detectable curvature.

The absence of nonlinear structure is itself a finding and closes the
robustness critique: collinearity does not distort cluster geometry
(Finding~I), the data naturally supports three clusters but a six-cluster
taxonomy is imposed (Finding~II), and all six kernel functions respect the
six-cluster assignment in 2D. The airline profitability data harbour no
detectable curvature.

\subsection{Finding VI: Kernel Ridge Regression Confirms Predictive Linearity}

Finding V established that no nonlinear \emph{structure} improves the
unsupervised geometry of the manifold. A complementary question is whether
nonlinear \emph{prediction} improves on the linear baseline: given the six
other 7D-raw variables, can operating profit be predicted more accurately by a
kernel model than by linear ridge regression? We fit kernel ridge regression
\cite{saunders1998} with RBF and degree-2 polynomial kernels against a linear
ridge baseline, using leave-one-out cross-validation (LOOCV) over the $n=25$
years with complete data (1995--2020; 2008 is absent from the original
data table reported in \cite{renold2023}). All six
predictors and the target were standardized within each training fold to
prevent test-fold leakage; kernel bandwidth ($\gamma$) and regularization
($\alpha$) were selected via a nested LOOCV grid search.

\begin{table}[ht]
 \centering
 \caption{Leave-one-out cross-validated predictive performance for
   Operating Profit, regressed on the remaining six 7D-raw variables
   ($n=25$, 1995--2020, excluding 2008).}
 \label{tab:krr}
 \small
 \begin{tabular}{@{}l r r@{}}
 \toprule
 \textbf{Model} & \textbf{$R^2$ (all years)} & \textbf{$R^2$ (excl.\ 2020)} \\
 \midrule
 Linear Ridge                 & 0.881 & 0.831 \\
 Kernel Ridge (RBF)           & 0.592 & 0.836 \\
 Kernel Ridge (poly., deg.\ 2) & 0.380 & 0.785 \\
 \bottomrule
 \end{tabular}
 \end{table}

Table~\ref{tab:krr} shows two regimes. Once the year 2020 is excluded, all
three models perform comparably ($R^2 = 0.79$--$0.84$): added kernel
flexibility buys no predictive advantage over the linear baseline,
corroborating Finding V's conclusion that the manifold is intrinsically linear
in-distribution.

But when 2020 is included, the kernel models collapse while linear ridge does
not.  Kernel ridge regression predicts from a weighted similarity to training
points; because 2020's load factor and RPM (a 65\% collapse) place it far
outside the training manifold, no training-fold observation is locally similar,
so the RBF and polynomial kernels revert toward the training-fold mean and
badly underpredict the shock.

This is a predictive-modeling echo of Findings I and V: the collapse is a known
limitation of similarity-based kernel extrapolation applied to a genuine,
isolated exogenous outlier, not evidence of unmodeled curvature in the
manifold.

\section{Discussion}
\label{sec:discussion}

This work also is a replication and re-analysis study: all techniques used are
established, and the contribution focuses on a systematic robustness check of a
published result rather than new methods.

The Fisher Discriminant Analysis \textsc{fda} requires pre-labelled membership;
the research question requires structure discovery, making FDA circular.
Statistically, with $n \approx 5$--$9$ per cluster, within-class covariance is
near-singular, destabilizing the computation. FDA would be useful post-hoc for
sharper discriminant directions. Labelled data would also enable supervised
manifold learning methods like linear discriminant analysis (LDA) or supervised
kernel PCA, which could further validate the linearity finding by checking if
supervised projections align with unsupervised ones.

The ARI = 1.0 finding (7D-raw vs 3D-PC) requires explanation.  Collinearity
stretches distance \emph{along} the shared growth direction without reordering
points \emph{within} it. This dataset has chronological gradient encoded in
\pc{1} (all variables grow monotonically 1995--2019 except crises). The
six-cluster taxonomy partitions this gradient into six contiguous bands.
Amplifying the growth axis approx. three times stretches all six bands
proportionally but does not reassign years across band boundaries, preserving
clusters.  This invariance would \emph{not} hold if clusters were orthogonal to
the collinear direction. 

Our replication demonstrates robustness from four angles: (1) cluster
assignments are invariant across spaces, explained by temporal alignment; (2)
3D geometric visualization reveals the temporal band structure; (3) kernel PCA
across six kernels spanning three families (distance-based,
information-geometric, graph-based) confirms the manifold is intrinsically
linear, with cross-kernel agreement on the \cluster{3}--\cluster{5} ambiguous
pair in 1D across all families; and (4) kernel ridge regression shows no
nonlinear advantage over linear ridge once 2020 is excluded, with the 2020
extrapolation failure attributable to local-similarity effects, not hidden
curvature. Together, these
findings validate Renold's headline claim that airline profitability regimes
are stable across spaces.

\section{Conclusion}
\label{sec:conclusion}

This study confirms the geometric robustness of Renold et al.'s clustering:
six-cluster assignments are invariant across 7D-raw and 3D-\pc{} space
(collinearity's growth direction is orthogonal to the temporal partition),
and six kernel PCA implementations confirm the manifold is linear, reinforced
by a kernel ridge regression check showing no nonlinear accuracy gain once
2020 is excluded. However, the silhouette criterion reveals a methodological
tension: the data structurally supports three clusters, not six --- a
suggestive, not confirmed, result (Finding II). The reversion-to-pre-COVID
claim itself is only assessed within this 1995--2020 window, since 2020 is
the terminal observation and no post-pandemic years exist to test it
directly.

On the methodological level this paper is attempting to stress two main
points: firstly, that a profound understanding of the data is key to any
machine learning approach; and secondly, it exemplifies the ground-breaking
work by Schölkopf and others on Kernel PCA almost three decades later.

\section{Future Work}

This study validated the airline manifold's linear character using kernel PCA
and classic methods. Several directions merit further investigation:

\emph{Temporal dynamics and regime switching.} A hidden Markov model or
regime-switching vector autoregression could quantify state persistence and
transition probabilities, moving beyond static clustering toward a dynamical
characterization.

\emph{Non-linear and modern representation-learning baselines.} This study
deliberately used only classical techniques for interpretability and
comparability with Renold et al. Broader techniques like autoencoders,
spectral/deep clustering, or contrastive representation learning, could test
whether the linear structure found here persists under representations that
don't assume linear separability.

\emph{Generalization to other transport sectors.} The participation-ratio
effect may apply to public transport data with similar secular trends; a
cross-sector study would test whether the collinearity–silhouette suppression
pattern is transport-specific or
domain-general.

\section{Acknowledgments}

The author thanks Prof.~Renold for fruitful discussions and for making the data
and methodology publicly accessible. To Prof.~Bunke I am forever thankful for
his mentorship and teaching at the FKI research group.

\bibliography{orthogonality-dimensionality}

\end{document}